\documentclass[10pt,twocolumn,letterpaper]{article}

\usepackage[pagenumbers]{cvpr} 

\usepackage{newfloat}
\usepackage{listings}
\usepackage{multirow}
\usepackage{makecell}
\usepackage{natbib}

\definecolor{cvprblue}{rgb}{0.21,0.49,0.74}
\usepackage[pagebackref,breaklinks,colorlinks,allcolors=cvprblue]{hyperref}

\def\paperID{} 
\def\confName{CVPR}
\def\confYear{2026}

\title{EmoCAST: Emotional Talking Portrait via Emotive Text Description}

\author{Yiguo Jiang$^{1}$ \ \ 
Xiaodong Cun$^{2}$\footnotemark[1] \ \ 
Yong Zhang$^{3}$ \ \ 
Yudian Zheng$^1$ \ \ 
Fan Tang$^4$ \ \ 
Chi-Man Pun$^{1}$\footnotemark[1] \ \ \\
\\
{$^1$University of Macau}  \quad  
{$^2$GVC Lab, Great Bay University}  \quad  
{$^3$Meituan}  \quad  
{$^4$ICT-CAS} \quad\\
\\
{Project Page: \url{https://github.com/GVCLab/EmoCAST}}
}

\begin{document}

\twocolumn[{
\renewcommand\twocolumn[1][]{#1}%
\maketitle
\begin{center}
    \vspace{-1.5em}
    \includegraphics[width=1.0\textwidth]{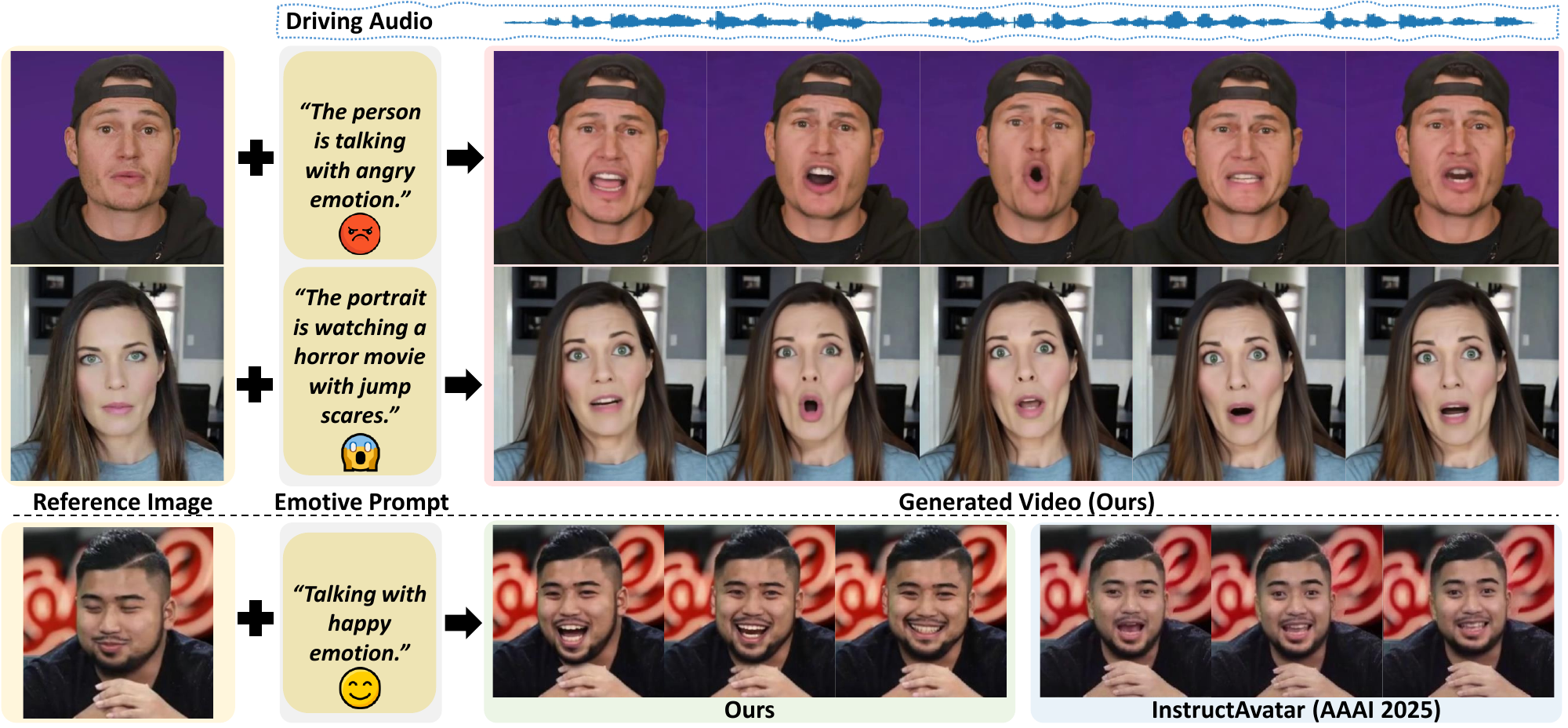}
    \captionof{figure}
    {We introduce EmoCAST, a novel diffusion-based emotional talking head system for \textit{in-the-wild} images that incorporates flexible and customizable emotive \textit{text prompts}.
     Compared with the previous state-of-the-art text-controlled emotional portrait animation method, \ie, InstructAvatar \cite{wang2024instructavatar}, EmoCAST produces more vivid and accurate facial expressions with better identity preservation.
    }
    \label{fig:teaser}
\end{center}
}]

\begin{abstract}

Emotional talking head synthesis aims to generate talking portrait videos with vivid expressions. 
Existing methods still exhibit limitations in control flexibility, motion naturalness, and expression quality. Moreover, currently available datasets are mainly collected in lab settings, further exacerbating these shortcomings and hindering real-world deployment.
To address these challenges, we propose EmoCAST, a diffusion-based talking head framework for precise, text-driven emotional synthesis. Its contributions are threefold: (1) architectural modules that enable effective text control; (2) an emotional talking-head dataset that expands the framework's ability; and (3) training strategies that further improve performance.
Specifically, for appearance modeling, emotional prompts are integrated through a text-guided emotive attention module, enhancing spatial knowledge to improve emotion understanding. 
To strengthen audio-emotion alignment, we introduce an emotive audio attention module to capture the interplay between controlled emotion and driving audio, generating emotion-aware features to guide precise facial motion synthesis. 
Additionally, we construct a large-scale, in-the-wild emotional talking head dataset with emotive text descriptions to optimize the framework's performance. Based on this dataset, we propose an emotion-aware sampling strategy and a progressive functional training strategy that improve the model's ability to capture nuanced expressive features and achieve accurate lip-sync. 
Overall, EmoCAST achieves state-of-the-art performance in generating realistic, emotionally expressive, and audio-synchronized talking-head videos.

\end{abstract}    
\vspace{-3mm}
\section{Introduction}
\label{sec:intro}

Generating vivid talking avatars has garnered significant attention in recent years. This technology offers diverse applications across multiple fields, including video content creation, animation production, digital humans, virtual reality, and human-machine interaction \cite{bozkurt2023speculative,zhou2020makelttalk,shen2023difftalk,tian2025emo}. 
Previous works \cite{prajwal2020lip,cheng2022videoretalking,guan2023stylesync,Zhang_2023_CVPR,stypulkowski2024diffused,xu2024hallo} have primarily concentrated on audio-lip synchronization in generated talking-head videos, ignoring the accompanying emotions, which is essential for natural human communication.

Some recent methods \cite{kaisiyuan2020mead,10.1145/3528233.3530745,Ma2023StyleTalkOT,wang2023progressive,Gan_2023_ICCV,zhong2024expclip,11045546,Wang_Guo_Bai_Yu_He_Tan_Sun_Bian_2025,ma2023dreamtalk,tan2025edtalk} have shifted their focus to emotion control in talking portrait generation, aiming to produce expressive and emotionally rich talking heads.
However, directly inferring expressions from speech remains challenging \cite{ma2024emobox}. For example, talking head videos generated using only emotionally cued audio often fail to exhibit distinct facial expressions \cite{Zhang_2023_CVPR,chen2025echomimic,xu2024hallo}. Consequently, additional emotion-control signals are typically required.
Early approaches utilize emotion labels to regulate expression categories in the generated talking videos~\cite{kaisiyuan2020mead,ji2021audio,Gan_2023_ICCV,xia2024gmtalker}, while others extract expression information directly from an emotional video template \cite{10.1145/3528233.3530745,Ma2023StyleTalkOT,wang2023progressive,ma2023dreamtalk}.
Nonetheless, these methods frequently encounter limitations in flexibility and controllability via the label or the reference video. 
Besides, since the emotional videos are hard to capture, existing emotional talking head generation datasets are still limited to the laboratory environment with restricted sample sizes and identities.

To address these challenges, we propose a novel diffusion-based framework for emotional talking head generation that leverages natural language for emotion control, thereby enhancing applicability to real-world scenarios.
We advance this goal along three axes: (i) design two modules that effectively integrate text control; (ii) construct an in-the-wild talking-head dataset with rich emotion annotations to facilitate accurate emotion modeling; and (iii) propose two training strategies to further optimize the framework.
Specifically, to achieve precise text-controlled emotional synthesis, our framework incorporates two key components: a text-guided emotive attention module and an emotive audio attention module. 
First, we design the text-guided emotive attention module to learn accurate alignment between emotional facial features and corresponding textual prompts in appearance modeling via a decoupled cross-attention mechanism. 
Beyond investigating the interaction between textual emotional features and facial features, the relationship between emotional features and audio signals requires systematic exploration. 
Accordingly, the emotive audio attention module aligns emotional information across textual emotion and audio modalities, modeling their correspondence for the facial region.

Furthermore, 
we construct a large-scale, in-the-wild Emotive Text-to-Talking Head (ETTH) dataset comprising 158 hours of emotional talking-head videos and spanning diverse identities. For each video, we provide accurate abstract emotion labels, fine-grained emotion intensity levels, and rich emotive textual descriptions.
Moreover, based on our dataset, we propose two training strategies. 
First, during expression learning training, instead of using the reference image from the same emotional video, we use a neutral-expression image of the same identity. This method significantly enhances the model's ability to capture subtle emotional nuances. 
Second, we propose a progressive functional training strategy that jointly leverages neutral and emotional talking-head datasets, progressively improving the model's generalization capacity, expression accuracy, and lip-synchronization in a coarse-to-fine manner.

To evaluate the effectiveness of our proposed method, we conduct comprehensive evaluations on both MEAD test set and in-the-wild test set.
The experimental results demonstrate that the proposed method achieves state-of-the-art performance in generating realistic, emotionally expressive talking-head videos. On the MEAD test set, our method attains an emotion accuracy of 83.60\%, substantially exceeding competing approaches. More importantly, on the out-of-domain, in-the-wild test set, it exhibits superior performance: both emotion accuracy and lip-sync quality surpass those of other methods, indicating strong generalization.

Overall, our main contributions are summarized as:
\begin{itemize}

\item We present EmoCAST, a novel framework for emotional talking portrait generation that integrates user-friendly emotional text prompts to produce lifelike expressions. 

\item To enable precise text-driven emotion control, we design two specific modules: a text-guided emotive attention module that aligns facial dynamics with textual prompts while preserving identity, and an emotive audio attention module to model the relationship between controlled emotion and driving speech.

\item We present a large-scale, in-the-wild emotional talking-head dataset with rich annotations, including discrete emotion categories, fine-grained emotion intensity levels, and textual emotion descriptions. We further propose two training strategies, namely emotion-aware sampling and progressive functional training.


\item Extensive experiments demonstrate that our method generates natural, emotionally expressive talking portraits that remain synchronized with the driving audio.

\end{itemize}

\section{Related Work}
\label{sec:related}

\begin{figure*}[t]
\centerline{\includegraphics[width=1 \linewidth]{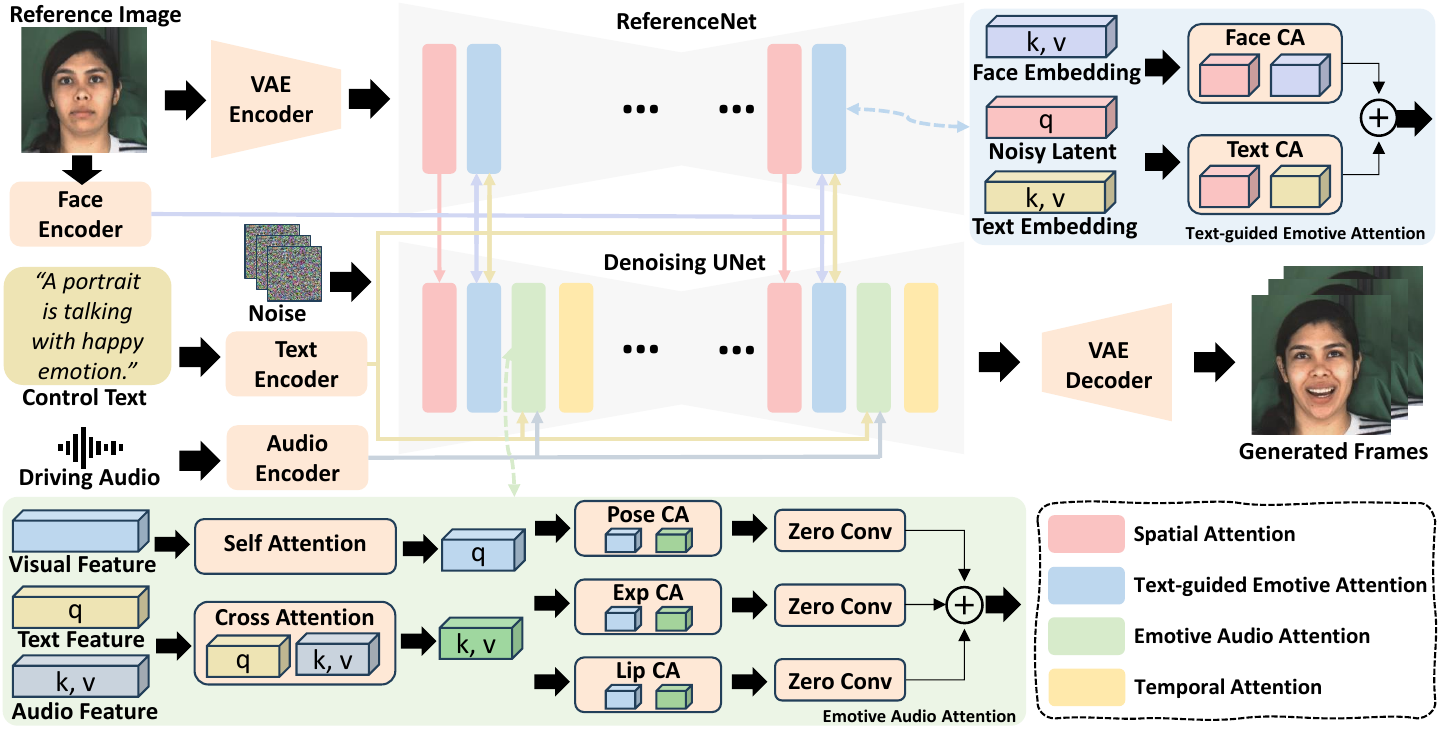}}
\caption{The main framework of the proposed EmoCAST, which has two pivotal modules designed for precise emotional synthesis: the \textbf{text-guided emotive attention module} and the \textbf{emotive audio attention module}. The text-guided emotive attention module ensures an accurate alignment between the generated facial expressions and the corresponding textual inputs. Concurrently, the emotive audio attention module facilitates the synthesis of facial motions that precisely reflect the emotional subtleties embedded in the driving speech.} 
\label{fig:method}
\end{figure*}

\noindent\textbf{Audio-Driven Talking Portrait Generation.}
Audio-driven talking portrait generation aims to create realistic talking head videos synchronized with corresponding speech.
Recently, some deep learning-based methods~\cite{zhou2020makelttalk,Zhang_2023_CVPR,Ma2023StyleTalkOT,wei2024aniportrait,chen2025echomimic,xu2024hallo,cui2024hallo2} have significantly advanced this domain.
MakeItTalk \cite{zhou2020makelttalk} predicts facial landmarks using disentangled audio content and speaker information. 
SadTalker \cite{Zhang_2023_CVPR} focuses on separately learning the expression and pose coefficients of a 3D Morphable Model.
EchoMimic \cite{chen2025echomimic} utilizes audio input and facial landmark to synthesize high-quality talking head videos.
Hallo \cite{xu2024hallo} employs a hierarchical audio-driven visual synthesis module to improve the precision of audio-visual alignment. 
Hallo2 \cite{cui2024hallo2} achieves long-duration, high-resolution portrait image animation.
The majority of these methods concentrate on generating synchronized mouth movements, neglecting the crucial aspect of emotional control.



\noindent\textbf{Emotional Audio-Driven Talking Portrait.}
Emotion significantly enhances the vividness and expressiveness of facial animation, thereby profoundly influencing the realism of generated talking portraits.
Recently, some audio-driven talking head methods have incorporated emotion control to produce more expressive and realistic talking portraits
\cite{kaisiyuan2020mead,10.1145/3528233.3530745,wang2023progressive,Gan_2023_ICCV,ma2023dreamtalk,11045546,Wang_Guo_Bai_Yu_He_Tan_Sun_Bian_2025,he2024emotalk3d,tan2025edtalk}.
EAMM~\cite{10.1145/3528233.3530745} extracts dynamic emotion patterns from a driven video and applies these transferable patterns to generate emotion-consistent talking heads. 
PD-FGC~\cite{wang2023progressive} employs disentangled latent representations to capture facial motion and subsequently inputs these latent into an image generator to synthesize talking heads. 
EAT~\cite{Gan_2023_ICCV} achieves emotion control through parameter-efficient adaptation of a pretrained emotion-agnostic talking head model.
EDTalk~\cite{tan2025edtalk} achieves effective emotion control by modeling expressions, mouth movements, and poses within three disentangled latent spaces.
TalkCLIP \cite{11045546} and InstructAvatar \cite{Wang_Guo_Bai_Yu_He_Tan_Sun_Bian_2025} rely on text-based control; however, generating accurate and vivid emotional expressions for in-the-wild reference images through textual control still remains challenging.
\section{Methodology}
\label{sec:method}

As shown in Fig.~\ref{fig:method}, given a single reference image as the appearance, the driving audio as talking content, and the text prompt for emotion modeling, the proposed EmoCAST generates expressive talking head videos with described emotion. 
Below, we first introduce the basic knowledge of the diffusion model in Sec.~\ref{sec:pre}, which establishes the foundational framework for our method. Sec.~\ref{sec:model} presents the EmoCAST pipeline with detailed explanations of its components. We then introduce our newly constructed Emotive Text-to-Talking Head (ETTH) dataset in Sec.~\ref{sec:data}. Finally, Sec.~\ref{sec:train} details the two proposed training strategies.


\subsection{Preliminaries: Latent Diffusion Model}
\label{sec:pre}

Diffusion Models~\cite{ho2020denoising,songdenoising}, especially the Latent Diffusion Model~(LDM)~\cite{rombach2022high}, produce data samples from Gaussian noise through iterative denoising steps. These models consist of two distinct phases: forward diffusion and backward denoising. During the forward diffusion process, Gaussian noise is progressively added to the original data. Conversely, the backward denoising process seeks to reconstruct the original data by reversing the noise addition procedure. 
We leverage the LDM for the talking head video generation task. 
Specifically, LDM utilizes the encoder $E$ of the pre-trained Variational Autoencoder~(VAE) to convert the input image $x$ to the latent space, generating initial latent feature $z_0=E(x)$. Subsequently, Gaussian noise $\epsilon \sim \mathcal{N}(\textbf{0}, \textbf{I})$ is gradually added to the latent feature $z_0$ over $t$ time steps, progressively diffusing towards the distribution $\mathcal{N}(\textbf{0}, \textbf{I})$. This diffusion process can be represented as:
$q(z_t|z_{t-1})=\mathcal{N}(z_t;\sqrt{1-\beta_t}z_{t-1},\beta_t \textbf{I}),$
where $\beta_t$ is a variance schedule. The $z_t$ in an arbitrary timestep $t$ of the diffusion process can be expressed as:
$q(z_t|z_0)=\mathcal{N}(z_t;\sqrt{\bar{\alpha}_t} z_0,(1 - \bar{\alpha}_t) \textbf{I}),$
where $\alpha_{t}=1-\beta_{t}$, $\bar{\alpha}_t=\prod_{s=1}^{t} a_s$.
Thus, $z_t$ can be derived from $z_0$, expressible as a linear combination of $z_0$ and the noise $\epsilon$ by
$z_t = \sqrt{\bar{\alpha}_t} z_0 + \sqrt{{1 - \bar{\alpha}_t}} \epsilon.$

During denoising, the UNet~\cite{ronneberger2015u} is trained to predict the added noise $\epsilon$ in the forward diffusion process. Consequently, the target latent $\hat{z_0}$ can be iteratively denoised from $z_t$. The objective function for training can be expressed as:
\begin{equation}\label{eq:diffusion-4}
\mathcal{L} = \text{E}_{z_t,\epsilon,c,t}[\|\epsilon-\epsilon_\theta(z_t,c,t)\|^2],
\end{equation}
where $\epsilon_\theta$ is the predicted noise by UNet, $c$ is condition set. After getting the target latent $\hat{z_0}$, the reconstructed output image $\hat{x}$ can be generated by a VAE decoder $\hat{x}=D(\hat{z_0})$. In our talking head animation task, we feed several latent features to the denoising network jointly for video modeling.

\subsection{Network Structure of EmoCAST}
\label{sec:model}

As shown in Fig.~\ref{fig:method}, our model primarily comprises ReferenceNet and Denoising UNet following the pre-trained Stable Diffusion~\cite{rombach2022high}, inspired by prior human animation methods~\cite{Hu_2024_CVPR, chen2025echomimic, xu2024hallo}.
The ReferenceNet extracts the visual appearance of the reference image and injects these features into Denoising UNet to guide frame generation. 
Denoising UNet progressively denoise noisy latents to produce emotional frames that maintain visual coherence with the reference image. 
Since our method is a video generation task, the temporal modules by temporal frame-wise attention~\cite {guo2023animatediff} are utilized to keep temporal consistency. 
Besides, audio is injected into the base model via cross-attention as motion control. 
Based on this network structure, we aim to generate an emotional talking portrait via the additional control text prompt. Thus, we propose a text-guided emotive attention module, which utilizes a decoupled cross-attention mechanism to feed the emotional textual feature into the diffusion model~(Sec.~\ref{sec:decoupled}).
Furthermore, 
we develop an emotive audio attention module to capture the relationship between emotive text and audio, thereby generating emotion-aware audio features to drive the synthesis of precise facial expression motions~(Sec.~\ref{sec:audio}).

\subsubsection{Text-guided Emotive Attention Module.}
\label{sec:decoupled}
As illustrated in Fig.~\ref{fig:method}, this module is designed to integrate face embeddings $e_f$ and text embeddings $e_t$ into the diffusion model. 
A straightforward approach is to concatenate textual embeddings $e_t$ and facial embeddings $e_f$, integrating them into the model through a shared cross-attention module. 
However, this method fails to effectively disentangle facial features from text-controlled attributes, often causing both the deterioration of identity-preserving visual features and insufficient learning of facial expressions from the control text.
To address this, we employ a decoupled cross-attention mechanism \cite{ye2023ip}, which more effectively captures expression features while preserving identity-related visual information.
Specifically, we first employ a pre-trained face encoder to extract facial embeddings $e_f$ for identity representation and utilize CLIP~\cite{radford2021learning} to obtain textual embeddings $e_t$ for emotion control.
Then, we utilize a decoupled cross-attention mechanism with two parallel branches: (1) Facial cross-attention $CA_{face}$ processes interactions between facial embeddings $e_f$ and noisy latent $z_t$. (2) Textual cross-attention $CA_{text}$ mediates interaction between textual embeddings $e_t$ and noisy latent $z_t$.
The final output combines both attention branches via addition:
\vspace{-4mm}
\begin{equation}
\begin{aligned}
\\&CA_{face}(Q(z_t),K(e_f),V(e_f))\\&\quad \ +CA_{text}(Q(z_t),K(e_t),V(e_t)) \label{eq:de}
\\&=Softmax(\frac{Q_z K_{f}^T}{\sqrt{d}}) V_f + Softmax(\frac{Q_z K_{t}^T}{\sqrt{d}}) V_t,
\end{aligned}
\end{equation}
where $Q_z= W_Q z_t$, $K_f=W_K^f e_f$, $V_f=W_V^f e_f$, $K_t=W_K^t e_t$, $V_t=W_V^t e_t$, and $W_Q$, $W_K^f$, $W_V^f$, $W_K^t$, $W_K^t$ are learnable projection matrices.
This design ensures that the generated facial features remain consistent with the reference image while simultaneously synthesizing vivid emotions that align with the provided emotional prompts.

\subsubsection{Emotive Audio Attention Module.}
\label{sec:audio}

To generate dynamic expression motions that are more consistent with emotional audio, we propose emotive audio attention module. This module first aligns audio features with textual emotion features to derive emotion-aware audio features, which are then used to interact with facial features, thereby guiding the generation of realistic dynamic facial expressions.
In detail, we first extract audio embedding using a pretrained wav2vec~\cite{schneider2019wav2vec}. 
For textual embeddings, we employ CLIP to provide emotional control information. Next, as shown in Fig. \ref{fig:method}, these extracted embeddings along with visual latent representation are jointly fed into the emotive audio attention module. 
To establish the relationship between textual expression features and audio features, the emotional text embedding $e_t$ undergoes a cross-attention operation with the audio embedding $e_a$ to obtain the emotion-aware audio feature $f_{ea}$. 
The calculation process is illustrated as follows:
\begin{equation}\label{eq:ea}
\begin{aligned}
f_{ea}=CA(Q(e_t),K(e_a),V(e_a)).
\end{aligned}
\end{equation}

Subsequently, the emotion-aware audio feature $f_{ea}$ and the visual latent feature $f_{v}$ are integrated through cross-attention to capture the relationships between audio and visual components.
Following Hallo \cite{xu2024hallo}, we implement three distinct cross-attention blocks for lips, expressions, and poses, respectively to extract corresponding features. 
The process is as follows:
\begin{equation}
\begin{aligned}
f_{lip}=CA(Q(f_v),K(f_{ea}),V(f_{ea}))\odot M_{lip},
\end{aligned}
\end{equation}
\begin{equation}
\begin{aligned}
f_{exp}=CA(Q(f_v),K(f_{ea}),V(f_{ea}))\odot M_{exp},
\end{aligned}
\end{equation}
\begin{equation}
\begin{aligned}
f_{pose}=CA(Q(f_v),K(f_{ea}),V(f_{ea}))\odot M_{pose},
\end{aligned}
\end{equation}
where $\odot$ is the Hadamard product. $M_{lip}$, $M_{exp}$, and $M_{pose}$ denote masks for the lip, expression, and pose regions, respectively. 
Finally, these features are combined using a convolutional layer and input to the subsequent module.



\begin{table}[t]
\caption{Comparison between ETTH and relevant datasets.} 
\begin{center}
\vspace{-4mm}
\resizebox{\linewidth}{!}{

\begin{tabular}{l | c c c c c}
\toprule[0.1em]
  &  IDs & Hours & Emo     & Emo    & Text\\ 
  &      &       & Label   & Level  & Description \\ 
\midrule[0.1em]

CelebV \cite{wayne2018reenactgan}& 5 & 2 & $\times$ & $\times$ & $\times$ \\
VoxCeleb \cite{nagrani2017voxceleb}& 1k+ & 352 & $\times$ & $\times$ & $\times$ \\
VoxCeleb2 \cite{chung2018voxceleb2}& 6k+ & \textbf{2442} & $\times$ & $\times$ & $\times$ \\
Hallo3 \cite{cui2024hallo3} & N/A & 70 & $\times$ & $\times$ & $\times$  \\
CelebV-HQ  \cite{zhu2022celebvhq}& \textbf{15k+} & 68 & $\checkmark$ & $\times$ & $\times$ \\
MEAD \cite{kaisiyuan2020mead}& 60 & 39 & $\checkmark$ & 3 & $\times $\\
EmoTalk3D \cite{he2024emotalk3d}& 30 & 15 & $\checkmark$ & 2 & $\times$ \\
\textbf{ETTH (Ours)} & \textbf{15k+} & 158 & $\checkmark$ & \textbf{Fine-grained} & $\checkmark$ \\

\bottomrule[0.1em]
\end{tabular}
}

\vspace{-4mm}
\label{table:etth}
\end{center}
\end{table}

\subsection{Emotive Text-to-Talking Head Dataset}
\label{sec:data}
The emotional talking head dataset is significantly smaller in scale compared to the extensive datasets of neutral talking head videos, as in Tab.~\ref{table:etth}. Furthermore, enabling fine-grained expression control via natural language necessitates datasets with detailed textual descriptions of emotional styles. To bridge these gaps, we introduce an Emotive Text-to-Talking Head~(ETTH) dataset featuring both accurate expression labels and rich emotive textual descriptions. 
Thus, we label the following datasets MEAD~\cite{kaisiyuan2020mead}, HDTF~\cite{Zhang2021FlowguidedOT}, CelebV-HQ~\cite{zhu2022celebvhq}, Hallo3~\cite{cui2024hallo3} from emotional aspects. 

In detail, we process the collected videos in three steps to meet our task requirements, including: lip synchronization filtering, emotion label annotation, and the generation of emotive text descriptions. 
For lip-sync, we use SyncNet~\cite{chung2017out} to obtain the Syn-C and Syn-D scores. This enables us to flexibly filter videos based on these metrics to meet diverse data requirements. Regarding emotion labels, we directly utilize the dataset-provided labels for the lab-collected MEAD videos. In the case of Hallo3 and CelebV-HQ, we employ Emotion-FAN~\cite{meng2019frame} that is fine-tuned on MEAD to generate abstract emotion labels and associated intensity values.
To generate emotional text prompts, we refer to MMHead \cite{wu2024mmhead} by providing ChatGPT with the video's abstract emotion label, enabling it to generate textual scene descriptions that evoke the target emotion.
The statistics of our ETTH dataset are detailed in Table \ref{table:etth}.
Our dataset encompasses a diverse range of speaker identities and includes comprehensive facial expression annotations. More details of the ETTH dataset are provided in the supplement.
\subsection{Progressive Emotion-aware Training}
\label{sec:train}

\begin{figure}[t]
\centerline{\includegraphics[width=\linewidth]{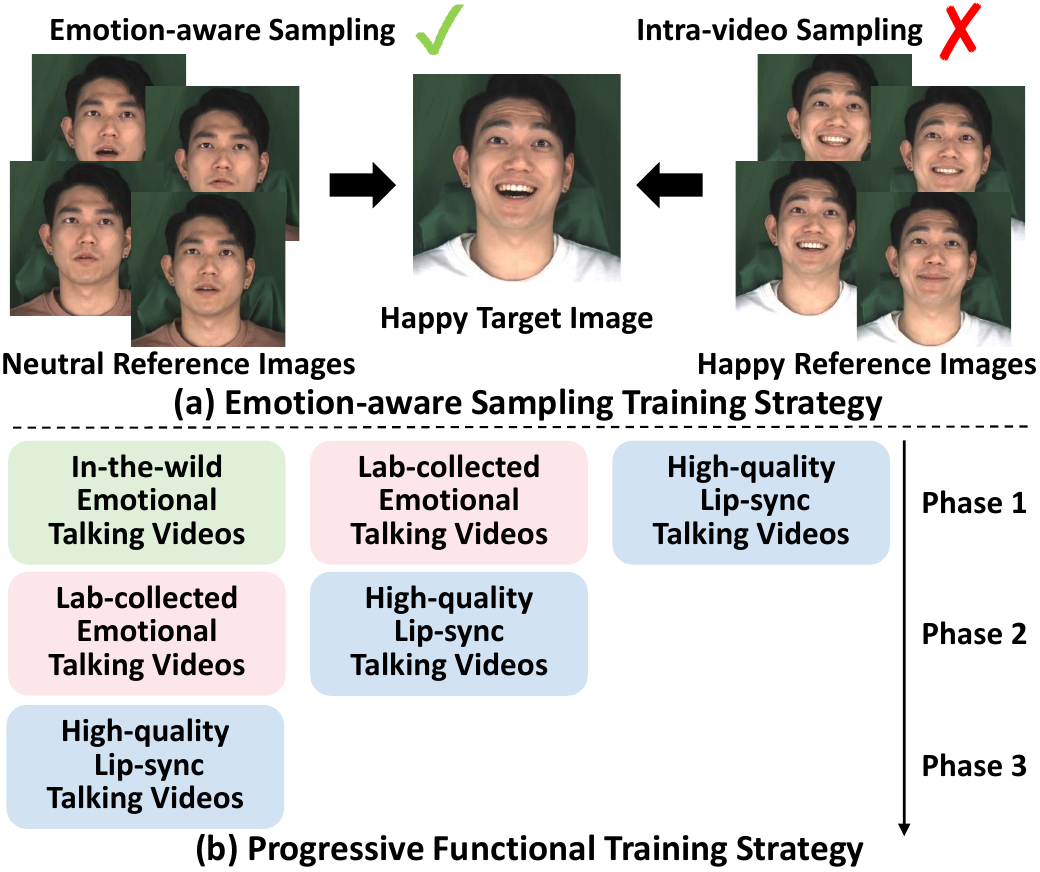}}
\caption{Visual illustration of the two proposed training strategies. (a)~Emotion-aware Sampling trains paired images between \textit{neutral} expression and \textit{emotional} expression to capture expression-specific features. (b)~Progressive Functional Training improves the model's generalization capability, expression accuracy, and lip-synchronization in a phased, coarse-to-fine manner.}
\vspace{-4mm}
\label{fig:train}

\end{figure}

Efficient use of our proposed dataset is critical for training high-performing emotional talking-head models. We demonstrate that training strategies are pivotal and introduce two key strategies.
First, we propose emotion-aware sampling strategy~(Sec.~\ref{sec:train1}), which enhances emotion modeling by learning the transformation from neutral to expressive facial representations. 
Second, we design a progressive functional training~(Sec.~\ref{sec:train2}), a coarse-to-fine scheme that hierarchically refines overall motion, emotional expression, and lip synchronization.
Concretely, our method initially trains the spatial layers to capture image-level expression information, enabling emotion-conditioned image-to-image generation. 
Then, the model is trained for temporal modeling. 
Building upon the learned emotional image generation, we implement a phased data sampling strategy to achieve audio-driven emotional video synthesis.

\begin{table*}[!t]
\centering
\caption{Quantitative comparisons with state-of-the-art methods on MEAD~\cite{kaisiyuan2020mead} and out-of-domain test sets. We mainly compare our method with diffusion-based methods, and the metrics of GAN-based methods are listed for reference. Best diffusion-based results are highlighted in bold.} 
\resizebox{\linewidth}{!}{
\begin{tabular}{l l l  c c c c  c c c }
\toprule[0.1em]
\multirow{3}{*}{Method} & \multirow{3}{*}{Backbone} & \multirow{3}{*}{\makecell{ Emotional \\ Condition}} & \multicolumn{4}{c}{MEAD Testset} & \multicolumn{3}{c}{ In-the-Wild Testset} \\ 
 \cmidrule(lr){4-7} \cmidrule(lr){8-10}
 & & & ${Acc}_{emo}\uparrow$ & LSE-D$\downarrow$ & LSE-C$\uparrow$ & \multicolumn{1}{c}{FID$\downarrow$} & ${Acc}_{emo}\uparrow$ & LSE-D$\downarrow$ & LSE-C$\uparrow$  \\ 
\midrule[0.1em]
MakeItTalk~\cite{zhou2020makelttalk} & GAN & N/A & 12.50\% & 9.78 & 5.25 & 73.92 & 12.86\% & 9.95 & 4.44  \\
SadTalker~\cite{Zhang_2023_CVPR} & GAN & N/A & 12.50\% & 7.49 & 7.60 & 62.79 & 12.50\% & 7.25 &  7.18 \\
EAMM~\cite{10.1145/3528233.3530745} & GAN & Video & 13.28\% & 11.11 & 3.96 & 76.70 & 21.79\% & 9.94 & 4.16 \\
PD-FGC~\cite{wang2023progressive} & GAN & Video & 43.75\% & 8.78 & 6.01 & 62.46 & 40.57\% & 9.18 &  5.19 \\
EDTalk~\cite{tan2025edtalk}  & GAN &  Video & 29.69\% & 7.17 & 8.06 & 59.60 & 33.57\% & 7.77 & 7.00  \\
EAT~\cite{Gan_2023_ICCV}  & GAN &  Label & 59.77\% & 7.69 & 7.91 & 58.21 & 32.50\% & 8.38 & 6.50  \\
\midrule[0.05em]
Aniportrait~\cite{wei2024aniportrait} & Diffusion & N/A & 12.50\% & 9.58 & 4.93 & 49.46 & 13.93\% & 10.35 &  3.72 \\ 
Echomimic~\cite{chen2025echomimic} & Diffusion & N/A & 12.50\% & 8.93 & 6.02 & 45.41 & 14.64\% & 9.13 & 5.49  \\ 
Hallo~\cite{xu2024hallo} & Diffusion & N/A & 12.50\% & 8.55 & 6.43 & 47.99 & 12.50\% & 8.34 &  6.23 \\ 
Hallo2~\cite{cui2024hallo2} & Diffusion & N/A & 12.50\% & \textbf{8.48} & 6.52 & 44.62 & 12.50\% & 8.39 & 6.19  \\ 
\textbf{Ours}  & Diffusion  & Text Prompt & \textbf{83.60\%} & 8.67 & \textbf{6.79} & \textbf{35.89} & \textbf{56.43\%} & \textbf{8.12} & \textbf{6.94}   \\
\bottomrule[0.1em]
\end{tabular}
}
\vspace{-1mm}
\label{table:compare}
\end{table*}

\subsubsection{Emotion-aware Sampling Training Strategy}
\label{sec:train1}
In the first training stage for emotional image-to-image generation, we employ an emotion-aware sampling strategy to enable effective learning of the distinctive characteristics of diverse emotional expressions.
Specifically, when training on a specific emotion, we avoid sampling both reference and target images from the same emotional video sequence. Instead, the target image is randomly sampled from the corresponding emotional video, while the reference image is randomly selected from the neutral expression video of the same identity as shown in Fig.\ref{fig:train}. This approach strengthens the model's ability to discern the differences between various expressions and neutral expressions, thereby improving its capacity to capture expression-specific features.

\subsubsection{Progressive Functional Training Strategy}
\label{sec:train2}

As illustrated in Fig. \ref{fig:train}, we introduce a progressive functional training strategy implemented in three phases:

\begin{figure*}[!t]
\centerline{\includegraphics[width=1 \linewidth]{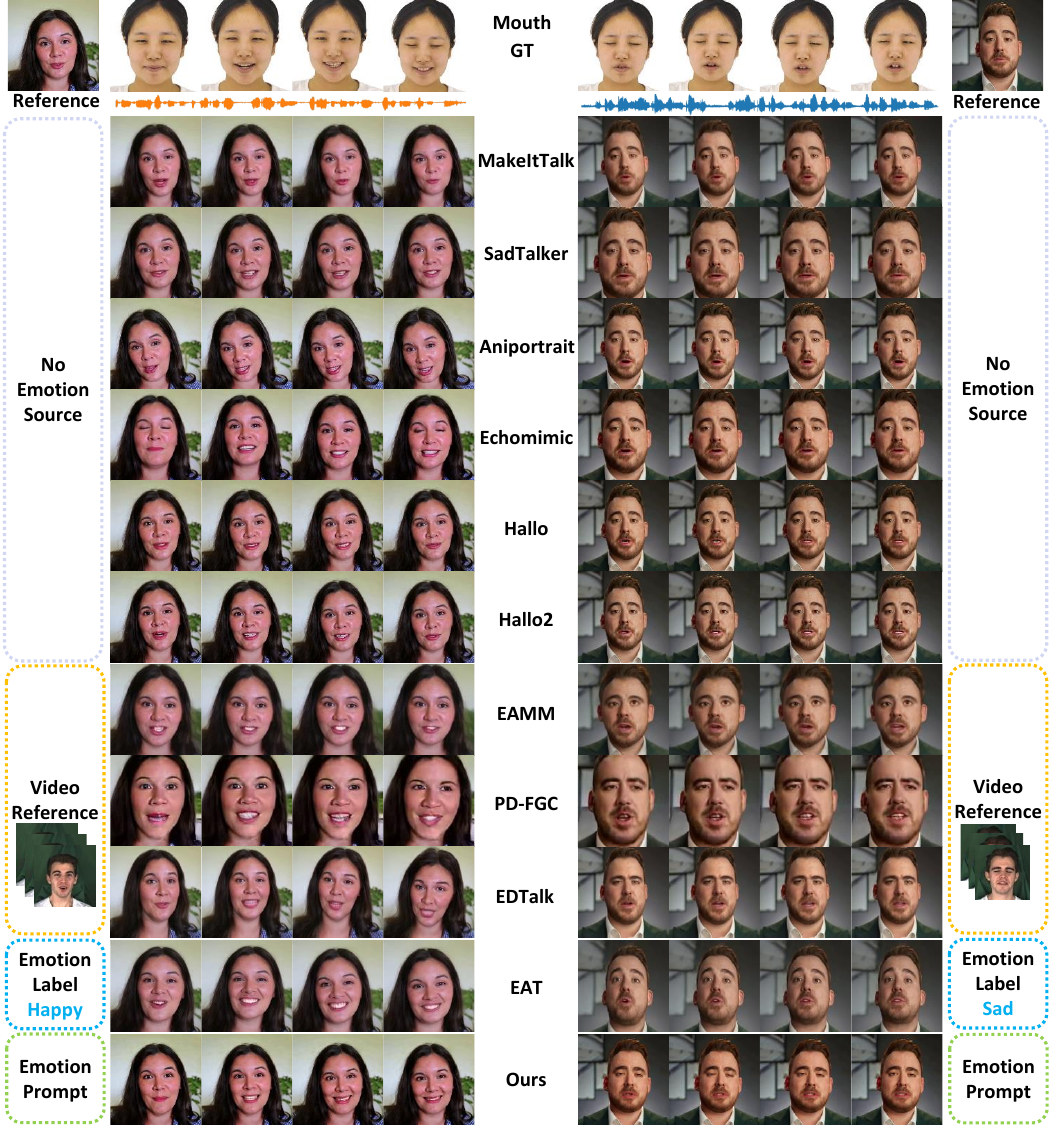}}
\caption{Visual comparison with other state-of-the-art methods for emotional talking video portraits on in-the-wild images. Our method consistently produces accurate facial expressions while maintaining precise lip synchronization that closely matches the ground truth mouth, along with robust identity preservation.
For a more detailed examination, kindly enlarge the image or view the supplemental video.} 
\label{fig:results}
\end{figure*}

\noindent\textbf{Phase 1 (Generalization Enhancement)}: First, we train the model on a mixed dataset including the in-the-wild emotional talking videos spanning diverse identities.
This phase enhances the model's generalization capability across diverse data sources.


\noindent\textbf{Phase 2 (Emotion Refinement)}: 
To refine facial expression accuracy and lip-sync, we exclude in-the-wild videos and train solely on a hybrid dataset comprising lab-collected emotional MEAD videos and high-quality lip-sync HDTF videos. This combination of two high-precision datasets ensures robustly generated results, even with limited identity.


\noindent\textbf{Phase 3 (Lip-Sync Specialization)}: Finally, to maximize lip-sync accuracy, we address potential interference from emotion by introducing an additional training phase. Specifically, we train the model on the HDTF, a high-quality talking-head dataset featuring neutral facial expressions and precise lip synchronization.


With this progressive functional training strategy, our model generates natural, emotionally expressive talking portraits with precise audio-visual synchronization.


\section{Experiments and Results}
\label{sec:exp}

\noindent\textbf{Dataset and Implementation Details.}
Our method is trained on an NVIDIA H800 GPU, using a batch size of 4 with 512 × 512 pixel videos.
For evaluation, following EAT \cite{Gan_2023_ICCV} and EDTalk \cite{tan2025edtalk}, we select four test subjects from MEAD \cite{kaisiyuan2020mead} and sample 256 emotional talking head videos, covering all 8 emotions.
To further assess generalization performance, we construct an additional in-the-wild out-of-domain test set comprising 7 reference images and 40 audio samples, resulting in 280 synthesized videos spanning 8 distinct emotional categories.

\noindent\textbf{Evaluation Metrics.}
To evaluate the generated emotional talking portrait videos, we employ several metrics. First, emotional accuracy of videos is assessed using the pre-trained emotion classifier~\cite{meng2019frame}, as referenced in EAT~\cite{Gan_2023_ICCV}. Second, audio-visual synchronization is measured using the lip-sync metrics~(LSE-D and LSE-C) from SyncNet~\cite{chung2017out}, as in Wav2Lip \cite{prajwal2020lip}. Finally, image quality of the synthesized portraits is evaluated using the Fréchet Inception Distance~(FID)~\cite{heusel2017gans}.

\noindent\textbf{Baselines.}
We perform a comparative analysis with state-of-the-art methods, including representative emotion-agnostic talking head approaches~(MakeItTalk~\cite{zhou2020makelttalk}, SadTalker~\cite{Zhang_2023_CVPR},
Aniportrait~\cite{wei2024aniportrait}, Echomimic~\cite{chen2025echomimic}, Hallo~\cite{xu2024hallo}, Hallo2~\cite{cui2024hallo2}) as well as open-source emotion-controllable talking head approaches~(EAMM~\cite{10.1145/3528233.3530745}, PD-FGC~\cite{wang2023progressive}, EAT~\cite{Gan_2023_ICCV}, and  EDTalk~\cite{tan2025edtalk}). 
For text-controlled methods TalkCLIP~\cite{11045546} and InstructAvatar~\cite{Wang_Guo_Bai_Yu_He_Tan_Sun_Bian_2025}, their source codes are not publicly available, making quantitative comparisons infeasible.
Accordingly, we extract reference images and driving audio from InstructAvatar’s official demo videos and use our method to generate talking-head videos for visual comparison.

\subsection{Comparison with Other Methods}\label{compare}
We perform quantitative comparisons with other methods on the MEAD test set \cite{kaisiyuan2020mead} and out-of-domain in-the-wild test set. 
Table \ref{table:compare} shows that our method outperforms competing approaches in emotional accuracy and visual quality, highlighting the effectiveness of our EmoCAST in achieving precise and vivid emotional representations. For audio-visual synchronization, our method performs comparably to existing techniques on the MEAD test set, while demonstrating superior performance on the in-the-wild test set, indicating stronger generalization. 

We further conduct visual comparisons with other state-of-the-art methods. 
As illustrated in Fig.~\ref{fig:results}, GAN-based methods exhibit lower visual fidelity and emotional expressiveness, leading to perceptibly unnatural emotional talking videos. Although EAMM~\cite{10.1145/3528233.3530745}, PD-FGC~\cite{wang2023progressive}, and  EDTalk~\cite{tan2025edtalk} utilize emotional videos as affective sources, their synthesized facial expressions remain insufficiently pronounced.
EAT~\cite{Gan_2023_ICCV} controls expression generation via emotional labels, enabling it to produce accurate expressions. However, the visual quality of these expressions is suboptimal, and the mouth sometimes fails to close in alignment with the ground truth. 
Moreover, as shown in Fig. \ref{fig:teaser}, under the same text-control setting, InstructAvatar \cite{wang2024instructavatar} yields weaker, less natural expressions and exhibits poor identity preservation.
In contrast, our approach achieves more vivid and faithful facial emotional details, maintains lip synchronization with the ground-truth lip movements, and robustly preserves identity.





\subsection{User Study}\label{userstudy}
\begin{table}[!t]
\begin{center}
\caption{User Study on In-the-Wild test set.}  
\vspace{-1mm}
\resizebox{\linewidth}{!}{
\begin{tabular}{l | c c c c c c c c}
\toprule[0.1em]
& SadTalker & Hallo2 & EAMM & PD-FGC  & EAT & Ours \\
\midrule[0.1em]
Audio-visual Sync & 3.13 & 3.30 & 1.29 & 2.31 & 3.11 & \textbf{3.68}  \\
Video Quality   & 3.23 & 3.63 & 1.20 & 1.34 & 2.66& \textbf{3.83} \\
Emotion Quality & 1.49 & 1.94 & 1.24 & 2.48 & 2.35  & \textbf{3.75} \\
\bottomrule[0.1em]
\end{tabular}
}
\vspace{-4mm}
\label{table:user}
\end{center}
\end{table}

To further evaluate the quality of the generated emotional talking portrait videos, we conduct a user study involving 22 participants. The study assesses the videos across three dimensions: emotion quality, audio-visual synchronization, and video quality, with scores ranging from 1 (minimum) to 5 (maximum). 
We compare 5 baseline methods with our proposed approach by sampling 10 videos from the in-the-wild test set, obtaining a total of 60 videos covering 8 emotions.
The results of user study are presented in Table \ref{table:user}. 





\begin{table}[!t]
\begin{center}
\caption{Quantitative results of the ablation study on MEAD.} 
\vspace{-1mm}
\resizebox{\linewidth}{!}{
\begin{tabular}{l | c c c c }
\toprule[0.1em]
  &  LSE-D $\downarrow$ & LSE-C $\uparrow$ & ${Acc}_{emo} \uparrow$ \\ 
\midrule[0.1em]
w/o text emotive attention & 8.91 & 6.55 & 44.92$\%$ \\
w/o emotive audio attention & 9.36 & 5.80 & 61.72$\%$ \\
w/o emotion-aware sampling & 8.82 & 6.57 &21.09$\%$ \\
w/o progressive training & 9.99 & 5.45 & 51.56$\%$ \\
Ours & \textbf{8.67} & \textbf{6.79} & \textbf{83.60$\%$} \\

\bottomrule[0.1em]
\end{tabular}
}
\vspace{-4mm}

\label{table:ab}
\end{center}
\end{table}

\begin{figure}[]
\centerline{\includegraphics[width=\linewidth]{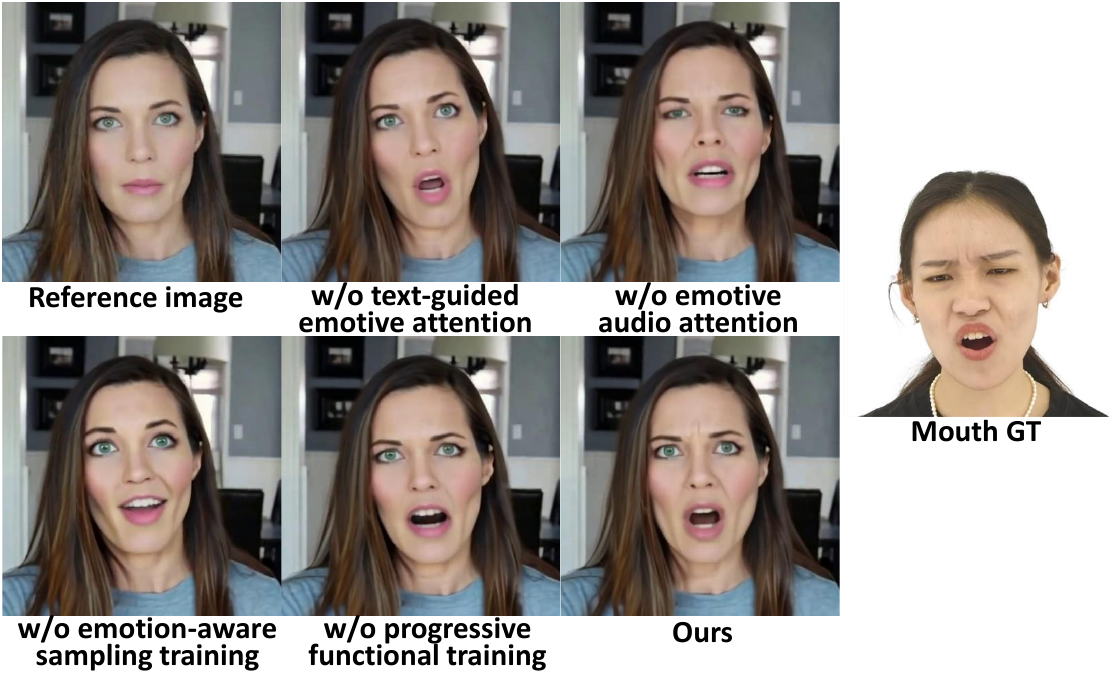}}
\vspace{-1mm}
\caption{Qualitative results of the ablation study for each design of our method. The emotion category is angry.}
\label{fig:ab}
\vspace{-3mm}
\end{figure}

\subsection{Ablation Studies}\label{abstudy}

We conduct comprehensive ablation studies to demonstrate the effectiveness of each design component. 

\noindent\textbf{Text-guided Emotive Attention Module.} 
We perform ablation studies to assess the text-guided emotive attention module’s capacity to learn appearance-level emotional cues.
We compare integrating emotional text and facial features through a shared cross-attention block with our decoupled emotive module.
The results are presented in Table \ref{table:ab} and Fig. \ref{fig:ab}. 
Relative to the shared cross-attention baseline, our text-guided decoupled module learns more precise expressions while better preserving identity.

\noindent\textbf{Emotive Audio Attention Module.} To evaluate the effectiveness of the interaction between speech and textual emotion, we conduct an ablation study by removing the interaction between textual emotion features and audio features in the emotive audio attention module. As illustrated in Fig.~\ref{fig:ab} and Table~\ref{table:ab}, enabling this interaction substantially improves performance, yielding more consistent facial motions that better synchronize with both the speech content and controlled expressions.

\noindent\textbf{Emotion-aware Sampling Training Strategy.} To validate the efficacy of our emotion-aware sampling training strategy, we compare it with the original intra-video sampling training mechanism, wherein both the reference image and the target image are selected from the same video.
As shown in Fig.~\ref{fig:ab} and Table~\ref{table:ab}, our emotion-aware sampling training strategy demonstrates the ability to learn more vivid and accurate expression information. 

\noindent\textbf{Progressive Functional Training Strategy.}
We conduct ablation studies to assess the progressive functional training strategy, which can generate highly natural and emotionally expressive talking portraits with precise audio-visual synchronization in a coarse-to-fine manner. 
For comparison, we evaluate a single-stage training baseline that uses all data simultaneously.
As shown in Table \ref{table:ab} and Fig. \ref{fig:ab}, the progressive training strategy produces more accurate facial expressions and significantly improves lip synchronization.

\section{Conclusion}
\label{sec:con}

We propose EmoCAST, a novel diffusion-based framework for generating customized, emotionally expressive talking head videos with flexible natural language for emotional control. The text prompts are efficiently integrated into the network via a text-guided emotive attention module and an emotive audio attention module, considering the relationships between emotion, appearance, and motion.
Furthermore, to address the scarcity of emotional datasets, we construct an Emotive Text-to-Talking Head~(ETTH) dataset containing precise expression labels and rich emotive textual descriptions. 
Based on this dataset, we propose an emotion-aware sampling strategy and a progressive functional training strategy, which further improve our model’s expression quality and lip-sync accuracy. 
Extensive experiments demonstrate that EmoCAST achieves state-of-the-art performance in generating highly natural and customizable expressive talking head videos.



{
    \small
    \bibliographystyle{ieeenat_fullname}
    \bibliography{main}

\begin{thebibliography}{47}
\providecommand{\natexlab}[1]{#1}
\providecommand{\url}[1]{\texttt{#1}}
\expandafter\ifx\csname urlstyle\endcsname\relax
  \providecommand{\doi}[1]{doi: #1}\else
  \providecommand{\doi}{doi: \begingroup \urlstyle{rm}\Url}\fi

\bibitem[Bozkurt et~al.(2023)Bozkurt, Junhong, Lambert, Pazurek, Crompton, Koseoglu, Farrow, Bond, Nerantzi, Honeychurch, et~al.]{bozkurt2023speculative}
Aras Bozkurt, Xiao Junhong, Sarah Lambert, Angelica Pazurek, Helen Crompton, Suzan Koseoglu, Robert Farrow, Melissa Bond, Chrissi Nerantzi, Sarah Honeychurch, et~al.
\newblock Speculative futures on chatgpt and generative artificial intelligence (ai): A collective reflection from the educational landscape.
\newblock \emph{Asian Journal of Distance Education}, 18\penalty0 (1):\penalty0 53--130, 2023.

\bibitem[Chen et~al.(2025)Chen, Cao, Chen, Li, and Ma]{chen2025echomimic}
Zhiyuan Chen, Jiajiong Cao, Zhiquan Chen, Yuming Li, and Chenguang Ma.
\newblock Echomimic: Lifelike audio-driven portrait animations through editable landmark conditions.
\newblock In \emph{Proceedings of the AAAI Conference on Artificial Intelligence}, pages 2403--2410, 2025.

\bibitem[Cheng et~al.(2022)Cheng, Cun, Zhang, Xia, Yin, Zhu, Wang, Wang, and Wang]{cheng2022videoretalking}
Kun Cheng, Xiaodong Cun, Yong Zhang, Menghan Xia, Fei Yin, Mingrui Zhu, Xuan Wang, Jue Wang, and Nannan Wang.
\newblock Videoretalking: Audio-based lip synchronization for talking head video editing in the wild.
\newblock In \emph{SIGGRAPH Asia 2022 Conference Papers}, pages 1--9, 2022.

\bibitem[Chung and Zisserman(2017)]{chung2017out}
Joon~Son Chung and Andrew Zisserman.
\newblock Out of time: automated lip sync in the wild.
\newblock In \emph{Computer Vision--ACCV 2016 Workshops: ACCV 2016 International Workshops, Taipei, Taiwan, November 20-24, 2016, Revised Selected Papers, Part II 13}, pages 251--263. Springer, 2017.

\bibitem[Chung et~al.(2018)Chung, Nagrani, and Zisserman]{chung2018voxceleb2}
Joon~Son Chung, Arsha Nagrani, and Andrew Zisserman.
\newblock Voxceleb2: Deep speaker recognition.
\newblock \emph{arXiv preprint arXiv:1806.05622}, 2018.

\bibitem[Cui et~al.(2024{\natexlab{a}})Cui, Li, Yao, Zhu, Shang, Cheng, Zhou, Zhu, and Wang]{cui2024hallo2}
Jiahao Cui, Hui Li, Yao Yao, Hao Zhu, Hanlin Shang, Kaihui Cheng, Hang Zhou, Siyu Zhu, and Jingdong Wang.
\newblock Hallo2: Long-duration and high-resolution audio-driven portrait image animation.
\newblock \emph{arXiv preprint arXiv:2410.07718}, 2024{\natexlab{a}}.

\bibitem[Cui et~al.(2024{\natexlab{b}})Cui, Li, Zhan, Shang, Cheng, Ma, Mu, Zhou, Wang, and Zhu]{cui2024hallo3}
Jiahao Cui, Hui Li, Yun Zhan, Hanlin Shang, Kaihui Cheng, Yuqi Ma, Shan Mu, Hang Zhou, Jingdong Wang, and Siyu Zhu.
\newblock Hallo3: Highly dynamic and realistic portrait image animation with video diffusion transformer.
\newblock \emph{arXiv preprint arXiv:2412.00733}, 2024{\natexlab{b}}.

\bibitem[Gan et~al.(2023)Gan, Yang, Yue, Sun, and Yang]{Gan_2023_ICCV}
Yuan Gan, Zongxin Yang, Xihang Yue, Lingyun Sun, and Yi Yang.
\newblock Efficient emotional adaptation for audio-driven talking-head generation.
\newblock In \emph{Proceedings of the IEEE/CVF International Conference on Computer Vision (ICCV)}, pages 22634--22645, 2023.

\bibitem[Guan et~al.(2023)Guan, Zhang, Zhou, HU, Wang, He, Feng, Liu, Ding, Liu, and Wang]{guan2023stylesync}
Jiazhi Guan, Zhanwang Zhang, Hang Zhou, Tianshu HU, Kaisiyuan Wang, Dongliang He, Haocheng Feng, Jingtuo Liu, Errui Ding, Ziwei Liu, and Jingdong Wang.
\newblock Stylesync: High-fidelity generalized and personalized lip sync in style-based generator.
\newblock In \emph{Proceedings of the IEEE/CVF Conference on Computer Vision and Pattern Recognition (CVPR)}, 2023.

\bibitem[Guo et~al.(2023)Guo, Yang, Rao, Liang, Wang, Qiao, Agrawala, Lin, and Dai]{guo2023animatediff}
Yuwei Guo, Ceyuan Yang, Anyi Rao, Zhengyang Liang, Yaohui Wang, Yu Qiao, Maneesh Agrawala, Dahua Lin, and Bo Dai.
\newblock Animatediff: Animate your personalized text-to-image diffusion models without specific tuning.
\newblock \emph{arXiv preprint arXiv:2307.04725}, 2023.

\bibitem[He et~al.(2024)He, Ji, Gong, Lu, Diao, Huang, Yao, Zhu, Ma, Xu, Wu, Zhang, Cao, and Zhu]{he2024emotalk3d}
Qianyun He, Xinya Ji, Yicheng Gong, Yuanxun Lu, Zhengyu Diao, Linjia Huang, Yao Yao, Siyu Zhu, Zhan Ma, Songchen Xu, Xiaofei Wu, Zixiao Zhang, Xun Cao, and Hao Zhu.
\newblock Emotalk3d: High-fidelity free-view synthesis of emotional 3d talking head.
\newblock In \emph{European Conference on Computer Vision (ECCV)}, 2024.

\bibitem[Heusel et~al.(2017)Heusel, Ramsauer, Unterthiner, Nessler, and Hochreiter]{heusel2017gans}
Martin Heusel, Hubert Ramsauer, Thomas Unterthiner, Bernhard Nessler, and Sepp Hochreiter.
\newblock Gans trained by a two time-scale update rule converge to a local nash equilibrium.
\newblock \emph{Advances in neural information processing systems}, 30, 2017.

\bibitem[Ho et~al.(2020)Ho, Jain, and Abbeel]{ho2020denoising}
Jonathan Ho, Ajay Jain, and Pieter Abbeel.
\newblock Denoising diffusion probabilistic models.
\newblock \emph{Advances in neural information processing systems}, 33:\penalty0 6840--6851, 2020.

\bibitem[Hu(2024)]{Hu_2024_CVPR}
Li Hu.
\newblock Animate anyone: Consistent and controllable image-to-video synthesis for character animation.
\newblock In \emph{Proceedings of the IEEE/CVF Conference on Computer Vision and Pattern Recognition (CVPR)}, pages 8153--8163, 2024.

\bibitem[Ji et~al.(2021)Ji, Zhou, Wang, Wu, Loy, Cao, and Xu]{ji2021audio}
Xinya Ji, Hang Zhou, Kaisiyuan Wang, Wayne Wu, Chen~Change Loy, Xun Cao, and Feng Xu.
\newblock Audio-driven emotional video portraits.
\newblock In \emph{Proceedings of the IEEE/CVF conference on computer vision and pattern recognition}, pages 14080--14089, 2021.

\bibitem[Ji et~al.(2022)Ji, Zhou, Wang, Wu, Wu, Xu, and Cao]{10.1145/3528233.3530745}
Xinya Ji, Hang Zhou, Kaisiyuan Wang, Qianyi Wu, Wayne Wu, Feng Xu, and Xun Cao.
\newblock Eamm: One-shot emotional talking face via audio-based emotion-aware motion model.
\newblock In \emph{ACM SIGGRAPH 2022 Conference Proceedings}, New York, NY, USA, 2022. Association for Computing Machinery.

\bibitem[Ma et~al.(2023{\natexlab{a}})Ma, Wang, Hu, Fan, Lv, Ding, Deng, and Yu]{Ma2023StyleTalkOT}
Yifeng Ma, Suzhe Wang, Zhipeng Hu, Changjie Fan, Tangjie Lv, Yu Ding, Zhidong Deng, and Xin Yu.
\newblock Styletalk: One-shot talking head generation with controllable speaking styles.
\newblock In \emph{AAAI Conference on Artificial Intelligence}, 2023{\natexlab{a}}.

\bibitem[Ma et~al.(2023{\natexlab{b}})Ma, Zhang, Wang, Wang, Zhang, and Deng]{ma2023dreamtalk}
Yifeng Ma, Shiwei Zhang, Jiayu Wang, Xiang Wang, Yingya Zhang, and Zhidong Deng.
\newblock Dreamtalk: When expressive talking head generation meets diffusion probabilistic models.
\newblock \emph{arXiv preprint arXiv:2312.09767}, 2023{\natexlab{b}}.

\bibitem[Ma et~al.(2025)Ma, Wang, Ding, Ma, Lv, Fan, Hu, Deng, and Yu]{11045546}
Yifeng Ma, Suzhen Wang, Yu Ding, Bowen Ma, Tangjie Lv, Changjie Fan, Zhipeng Hu, Zhidong Deng, and Xin Yu.
\newblock Talkclip: Talking head generation with text-guided expressive speaking styles.
\newblock \emph{IEEE Transactions on Multimedia}, pages 1--12, 2025.

\bibitem[Ma et~al.(2024)Ma, Chen, Zhang, Zheng, Chen, Li, Ye, Chen, and Hain]{ma2024emobox}
Ziyang Ma, Mingjie Chen, Hezhao Zhang, Zhisheng Zheng, Wenxi Chen, Xiquan Li, Jiaxin Ye, Xie Chen, and Thomas Hain.
\newblock Emobox: Multilingual multi-corpus speech emotion recognition toolkit and benchmark.
\newblock In \emph{Proc. INTERSPEECH}, 2024.

\bibitem[Meng et~al.(2019)Meng, Peng, Wang, and Qiao]{meng2019frame}
Debin Meng, Xiaojiang Peng, Kai Wang, and Yu Qiao.
\newblock Frame attention networks for facial expression recognition in videos.
\newblock In \emph{2019 IEEE international conference on image processing (ICIP)}, pages 3866--3870. IEEE, 2019.

\bibitem[Nagrani et~al.(2017)Nagrani, Chung, and Zisserman]{nagrani2017voxceleb}
Arsha Nagrani, Joon~Son Chung, and Andrew Zisserman.
\newblock Voxceleb: a large-scale speaker identification dataset.
\newblock \emph{arXiv preprint arXiv:1706.08612}, 2017.

\bibitem[Prajwal et~al.(2020)Prajwal, Mukhopadhyay, Namboodiri, and Jawahar]{prajwal2020lip}
KR Prajwal, Rudrabha Mukhopadhyay, Vinay~P Namboodiri, and CV Jawahar.
\newblock A lip sync expert is all you need for speech to lip generation in the wild.
\newblock In \emph{Proceedings of the 28th ACM international conference on multimedia}, pages 484--492, 2020.

\bibitem[Radford et~al.(2021)Radford, Kim, Hallacy, Ramesh, Goh, Agarwal, Sastry, Askell, Mishkin, Clark, et~al.]{radford2021learning}
Alec Radford, Jong~Wook Kim, Chris Hallacy, Aditya Ramesh, Gabriel Goh, Sandhini Agarwal, Girish Sastry, Amanda Askell, Pamela Mishkin, Jack Clark, et~al.
\newblock Learning transferable visual models from natural language supervision.
\newblock In \emph{International conference on machine learning}, pages 8748--8763. PmLR, 2021.

\bibitem[Rombach et~al.(2022)Rombach, Blattmann, Lorenz, Esser, and Ommer]{rombach2022high}
Robin Rombach, Andreas Blattmann, Dominik Lorenz, Patrick Esser, and Bj{\"o}rn Ommer.
\newblock High-resolution image synthesis with latent diffusion models.
\newblock In \emph{Proceedings of the IEEE/CVF conference on computer vision and pattern recognition}, pages 10684--10695, 2022.

\bibitem[Ronneberger et~al.(2015)Ronneberger, Fischer, and Brox]{ronneberger2015u}
Olaf Ronneberger, Philipp Fischer, and Thomas Brox.
\newblock U-net: Convolutional networks for biomedical image segmentation.
\newblock In \emph{Medical image computing and computer-assisted intervention--MICCAI 2015: 18th international conference, Munich, Germany, October 5-9, 2015, proceedings, part III 18}, pages 234--241. Springer, 2015.

\bibitem[Schneider et~al.(2019)Schneider, Baevski, Collobert, and Auli]{schneider2019wav2vec}
Steffen Schneider, Alexei Baevski, Ronan Collobert, and Michael Auli.
\newblock wav2vec: Unsupervised pre-training for speech recognition.
\newblock \emph{arXiv preprint arXiv:1904.05862}, 2019.

\bibitem[Shen et~al.(2023)Shen, Zhao, Meng, Li, Zhu, Zhou, and Lu]{shen2023difftalk}
Shuai Shen, Wenliang Zhao, Zibin Meng, Wanhua Li, Zheng Zhu, Jie Zhou, and Jiwen Lu.
\newblock Difftalk: Crafting diffusion models for generalized audio-driven portraits animation.
\newblock In \emph{CVPR}, 2023.

\bibitem[Song et~al.(2021)Song, Meng, and Ermon]{songdenoising}
Jiaming Song, Chenlin Meng, and Stefano Ermon.
\newblock Denoising diffusion implicit models.
\newblock In \emph{International Conference on Learning Representations}, 2021.

\bibitem[Stypu{l}kowski et~al.(2024)Stypu{l}kowski, Vougioukas, He, Zi{k{e}}ba, Petridis, and Pantic]{stypulkowski2024diffused}
Micha{l} Stypu{l}kowski, Konstantinos Vougioukas, Sen He, Maciej Zi{k{e}}ba, Stavros Petridis, and Maja Pantic.
\newblock Diffused heads: Diffusion models beat gans on talking-face generation.
\newblock In \emph{Proceedings of the IEEE/CVF Winter Conference on Applications of Computer Vision}, pages 5091--5100, 2024.

\bibitem[Tan et~al.(2025)Tan, Ji, Bi, and Pan]{tan2025edtalk}
Shuai Tan, Bin Ji, Mengxiao Bi, and Ye Pan.
\newblock Edtalk: Efficient disentanglement for emotional talking head synthesis.
\newblock In \emph{European Conference on Computer Vision}, pages 398--416. Springer, 2025.

\bibitem[Tian et~al.(2025)Tian, Wang, Zhang, and Bo]{tian2025emo}
Linrui Tian, Qi Wang, Bang Zhang, and Liefeng Bo.
\newblock Emo: Emote portrait alive generating expressive portrait videos with audio2video diffusion model under weak conditions.
\newblock In \emph{European Conference on Computer Vision}, pages 244--260. Springer, 2025.

\bibitem[Wang et~al.(2023)Wang, Deng, Yin, Shum, and Wang]{wang2023progressive}
Duomin Wang, Yu Deng, Zixin Yin, Heung-Yeung Shum, and Baoyuan Wang.
\newblock Progressive disentangled representation learning for fine-grained controllable talking head synthesis.
\newblock In \emph{Proceedings of the IEEE/CVF Conference on Computer Vision and Pattern Recognition (CVPR)}, pages 17979--17989, 2023.

\bibitem[Wang et~al.(2020)Wang, Wu, Song, Yang, Wu, Qian, He, Qiao, and Loy]{kaisiyuan2020mead}
Kaisiyuan Wang, Qianyi Wu, Linsen Song, Zhuoqian Yang, Wayne Wu, Chen Qian, Ran He, Yu Qiao, and Chen~Change Loy.
\newblock Mead: A large-scale audio-visual dataset for emotional talking-face generation.
\newblock In \emph{ECCV}, 2020.

\bibitem[Wang et~al.(2024)Wang, Guo, Bai, Yu, He, Tan, Sun, and Bian]{wang2024instructavatar}
Yuchi Wang, Junliang Guo, Jianhong Bai, Runyi Yu, Tianyu He, Xu Tan, Xu Sun, and Jiang Bian.
\newblock Instructavatar: Text-guided emotion and motion control for avatar generation.
\newblock \emph{arXiv preprint arXiv:2405.15758}, 2024.

\bibitem[Wang et~al.(2025)Wang, Guo, Bai, Yu, He, Tan, Sun, and Bian]{Wang_Guo_Bai_Yu_He_Tan_Sun_Bian_2025}
Yuchi Wang, Junliang Guo, Jianhong Bai, Runyi Yu, Tianyu He, Xu Tan, Xu Sun, and Jiang Bian.
\newblock Instructavatar: Text-guided emotion and motion control for avatar generation.
\newblock \emph{Proceedings of the AAAI Conference on Artificial Intelligence}, pages 8132--8140, 2025.

\bibitem[Wei et~al.(2024)Wei, Yang, and Wang]{wei2024aniportrait}
Huawei Wei, Zejun Yang, and Zhisheng Wang.
\newblock Aniportrait: Audio-driven synthesis of photorealistic portrait animation.
\newblock \emph{arXiv preprint arXiv:2403.17694}, 2024.

\bibitem[Wu et~al.(2024)Wu, Li, Yan, Duan, Liu, and Zhai]{wu2024mmhead}
Sijing Wu, Yunhao Li, Yichao Yan, Huiyu Duan, Ziwei Liu, and Guangtao Zhai.
\newblock Mmhead: Towards fine-grained multi-modal 3d facial animation.
\newblock In \emph{Proceedings of the 32nd ACM International Conference on Multimedia}, pages 7966--7975, 2024.

\bibitem[Wu et~al.(2018)Wu, Zhang, Li, Qian, and Loy]{wayne2018reenactgan}
Wayne Wu, Yunxuan Zhang, Cheng Li, Chen Qian, and Chen~Change Loy.
\newblock Reenactgan: Learning to reenact faces via boundary transfer.
\newblock In \emph{ECCV}, 2018.

\bibitem[Xia et~al.(2024)Xia, Wang, Deng, Luo, and Liu]{xia2024gmtalker}
Yibo Xia, Lizhen Wang, Xiang Deng, Xiaoyan Luo, and Yebin Liu.
\newblock Gmtalker: Gaussian mixture-based audio-driven emotional talking video portraits.
\newblock \emph{arXiv preprint arXiv:2312.07669v2}, 2024.

\bibitem[Xu et~al.(2024)Xu, Li, Su, Shang, Zhang, Liu, Wang, Yao, and zhu]{xu2024hallo}
Mingwang Xu, Hui Li, Qingkun Su, Hanlin Shang, Liwei Zhang, Ce Liu, Jingdong Wang, Yao Yao, and Siyu zhu.
\newblock Hallo: Hierarchical audio-driven visual synthesis for portrait image animation, 2024.

\bibitem[Ye et~al.(2023)Ye, Zhang, Liu, Han, and Yang]{ye2023ip}
Hu Ye, Jun Zhang, Sibo Liu, Xiao Han, and Wei Yang.
\newblock Ip-adapter: Text compatible image prompt adapter for text-to-image diffusion models.
\newblock \emph{arXiv preprint arXiv:2308.06721}, 2023.

\bibitem[Zhang et~al.(2023)Zhang, Cun, Wang, Zhang, Shen, Guo, Shan, and Wang]{Zhang_2023_CVPR}
Wenxuan Zhang, Xiaodong Cun, Xuan Wang, Yong Zhang, Xi Shen, Yu Guo, Ying Shan, and Fei Wang.
\newblock Sadtalker: Learning realistic 3d motion coefficients for stylized audio-driven single image talking face animation.
\newblock In \emph{Proceedings of the IEEE/CVF Conference on Computer Vision and Pattern Recognition (CVPR)}, pages 8652--8661, 2023.

\bibitem[Zhang et~al.(2021)Zhang, Li, and Ding]{Zhang2021FlowguidedOT}
Zhimeng Zhang, Lincheng Li, and Yu Ding.
\newblock Flow-guided one-shot talking face generation with a high-resolution audio-visual dataset.
\newblock \emph{2021 IEEE/CVF Conference on Computer Vision and Pattern Recognition (CVPR)}, pages 3660--3669, 2021.

\bibitem[Zhong et~al.(2024)Zhong, Wei, Yang, and Wang]{zhong2024expclip}
Yicheng Zhong, Huawei Wei, Peiji Yang, and Zhisheng Wang.
\newblock Expclip: Bridging text and facial expressions via semantic alignment.
\newblock In \emph{Proceedings of the AAAI Conference on Artificial Intelligence}, pages 7614--7622, 2024.

\bibitem[Zhou et~al.(2020)Zhou, Han, Shechtman, Echevarria, Kalogerakis, and Li]{zhou2020makelttalk}
Yang Zhou, Xintong Han, Eli Shechtman, Jose Echevarria, Evangelos Kalogerakis, and Dingzeyu Li.
\newblock Makelttalk: speaker-aware talking-head animation.
\newblock \emph{ACM Transactions On Graphics (TOG)}, 39\penalty0 (6):\penalty0 1--15, 2020.

\bibitem[Zhu et~al.(2022)Zhu, Wu, Zhu, Jiang, Tang, Zhang, Liu, and Loy]{zhu2022celebvhq}
Hao Zhu, Wayne Wu, Wentao Zhu, Liming Jiang, Siwei Tang, Li Zhang, Ziwei Liu, and Chen~Change Loy.
\newblock {CelebV-HQ}: A large-scale video facial attributes dataset.
\newblock In \emph{ECCV}, 2022.

\end{thebibliography}
}


\end{document}